\documentclass{article}

\usepackage{arxiv}

\usepackage[utf8]{inputenc} 
\usepackage[T1]{fontenc}    
\usepackage{hyperref}       
\usepackage{url}            
\usepackage{booktabs}       
\usepackage{amsfonts}       
\usepackage{nicefrac}       
\usepackage{microtype}      
\usepackage{lipsum}		
\usepackage{graphicx}
\usepackage{natbib}
\usepackage{doi}

\usepackage{amsthm}
\usepackage{floatrow}
\usepackage{multirow}
\usepackage[misc]{ifsym}
\usepackage{float}
\floatstyle{plaintop}
\restylefloat{table}
\usepackage{siunitx}
\usepackage{tabularx}
\usepackage{subcaption}
\usepackage[english]{babel}
\newtheorem{theorem}{Theorem}
\usepackage{amsmath}
\newtheorem{definition}{Definition}[section]

\title{Transforming PageRank into an Infinite-Depth Graph Neural Network}

\author{ \href{https://orcid.org/0000-0002-0515-7635}{\includegraphics[scale=0.06]{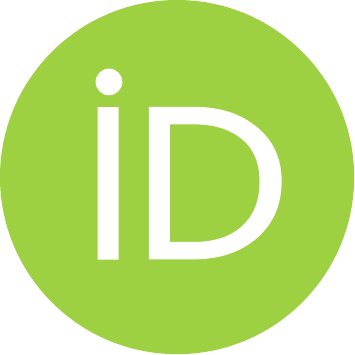}\hspace{1mm}Andreas Roth} \\
	Artificial Intelligence Group\\
	TU Dortmund\\
	Dortmund, Germany \\
	\texttt{andreas.roth@tu-dortmund.de} \\
	\And
	\href{https://orcid.org/0000-0002-9841-1101}{\includegraphics[scale=0.06]{orcid.pdf}\hspace{1mm}Thomas Liebig} \\
	Artificial Intelligence Group\\
	TU Dortmund\\
	Dortmund, Germany \\
	\texttt{thomas.liebig@tu-dortmund.de} \\
}

\renewcommand{\headeright}{Accepted at ECML-PKDD 2022}
\renewcommand{\undertitle}{Accepted at ECML-PKDD 2022}
\renewcommand{\shorttitle}{Personalized PageRank Graph Neural Network}

\hypersetup{
pdftitle={Transforming PageRank into an Infinite-Depth Graph Neural Network},
pdfsubject={Machine Learning, Graph Neural Networks, PageRank},
pdfauthor={Andreas Roth, Thomas Liebig},
pdfkeywords={Machine Learning, Graph Neural Networks, PageRank},
}

\begin{document}
\maketitle

\begin{abstract}
Popular graph neural networks are shallow models, despite the success of very deep architectures in other application domains of deep learning.
This reduces the modeling capacity and leaves models unable to capture long-range relationships.
The primary reason for the shallow design results from over-smoothing, which leads node states to become more similar with increased depth.
We build on the close connection between GNNs and PageRank, for which personalized PageRank introduces the consideration of a personalization vector.
Adopting this idea, we propose the Personalized PageRank Graph Neural Network (PPRGNN), which extends the graph convolutional network to an infinite-depth model that has a chance to reset the neighbor aggregation back to the initial state in each iteration.
We introduce a nicely interpretable tweak to the chance of resetting and prove the convergence of our approach to a unique solution without placing any constraints, even when taking infinitely many neighbor aggregations. As in personalized PageRank, our result does not suffer from over-smoothing.
While doing so, time complexity remains linear while we keep memory complexity constant, independently of the depth of the network, making it scale well to large graphs. 
We empirically show the effectiveness of our approach for various node and graph classification tasks. PPRGNN outperforms comparable methods in almost all cases.
\footnote{Our code is available at: https://github.com/roth-andreas/pprgnn}
\end{abstract}

\keywords{Machine Learning \and Graph Neural Networks  \and PageRank}

\section{Introduction}
Graph-structured data is found in many real-world applications ranging from social networks~\citep{otte2002social} to biological structures~\citep{pavlopoulos2011using}. Steadily growing amounts of data lead to emerging solutions that can extract relevant information from these data types. 
Tasks like providing recommendations~\citep{ying2018graph}, predicting the state of traffic~\citep{derrow2021eta} or the classification of entire graphs into distinct categories~\citep{yanardag2015deep} are some of the tasks of research interest.
Approaches based on deep learning have found great success for grid-structured data, e.g., in image processing~\citep{he2016deep} and natural language processing~\citep{vaswani2017attention}. 
Graph Neural Networks (GNNs)~\citep{kipf2016semi} adopt the ideas from convolutions in euclidean space for irregular non-euclidean domains.
These methods directly consider the graph structure when performing convolution operations.

One of the challenges of GNNs is to capture long-range dependencies. 
Recently popular methods use an aggregation scheme, in which $k$ layers of graph convolutions combine the information from $k$-hops around each node~\citep{kipf2016semi,velickovic2018graph}. 
Several issues, most dominantly over-smoothing~\citep{li2018deeper,xu2018representation} and memory consumption~\citep{hamilton2017inductive,chiang2019cluster,zeng2020graphsaint} were found to prevent deep models, as in image processing~\citep{he2016deep}.
Several recent efforts explore options to enable more layers and even formulate infinite-depth equations. However, previous work still only allows a limited depth~\citep{rong2020dropedge,xu2018representation} or places hard constraints on the parameters~\citep{gu2020implicit} or the architecture~\citep{bai2019deep}.

As identified by~\citet{klicpera2018predict}, GNNs are closely related to PageRank~\citep{page1999pagerank}, which in its basic version only depends on the graph structure, not on the initial distribution. Personalized PageRank~\citep{page1999pagerank} introduces a chance to reset PageRank to a teleportation vector, allowing the result to depend not only on the graph structure but also on an initial distribution. We show how the idea of personalization can be adopted to GNNs and propose the Personalized PageRank Graph Neural Network (PPRGNN), an infinitely deep GNN that adds a chance to reset the neighbor aggregation back to the initial state. In order to prove the convergence of PPRGNN to a unique solution when iterating infinitely many times, we modify the chance of resetting to be dynamic based on the distance to the root node. As in personalized PageRank, our approach does not suffer from over-smoothing and the locality of node features around their root nodes is preserved. Due to the large depths, far distant information still influences resulting node representations.

In addition, we provide rich theoretical intuition for the success of our formulation and our design choices.
While the depth is theoretically always infinite, the practically effective depth is adaptive and varies depending on the learned parameters, the graph structure, and the observed features.
We also provide a way to control the convergence rate since different levels of localization are effective for different types of graphs~\citep{abu2018watch,grover2016node2vec}.
Furthermore, contrary to previous infinite-depth approaches, we do not impose any constraints on parameters or the model's architecture.
To allow scaling to large graphs despite the infinite depth, we design an efficient gradient computation that remains constant in memory and execution time.
We validate our proposed approach against comparable methods on various inductive and transductive node and graph classification tasks. Our approach outperforms related methods in almost all cases by considerable margins, while most other approaches are within a competitive range only for individual tasks. The experimental execution time is also improved compared to previous infinite-depth approaches.    

The rest of our work is structured as follows. Section~\ref{sec:preliminaries} introduces our notation and relevant basics in personalized PageRank and GNNs. We describe recent related approaches in Section~\ref{sec:related}. Our method is detailed in Section~\ref{sec:methods}, and a comprehensive evaluation is presented in Section~\ref{sec:experiments}. We discuss our results and potential future directions in Section~\ref{sec:conclusion}.  
\section{Preliminaries}
\label{sec:preliminaries}
We represent a graph $G = (V, E)$ as the tuple of $n$ nodes $V = \{v_1,v_2,\dots,v_n\}$ and a set of edges $E$ between pairs of nodes. We construct an adjacency matrix $\mathbf{A} \in \mathbb{R}^{n \times n}$ describing the connectivity between pairs of nodes from the $E$. Entries $a_{ij} \in \mathbf{A}$ indicate the strength of an edge between nodes $v_i$ and $v_j$, a zero-entry indicates the absence of an edge. Our method assumes undirected edges, e.g., $a_{ij} = a_{ji}$, but it is straightforward to apply it to directed graphs. We use a normalized version $\mathbf{\Tilde{A}} = \mathbf{D}^{-1/2}\mathbf{A}\mathbf{D}^{-1/2}$ of the adjacency matrix, potentially with self-loops.
Each node $v_k$ has a set of $F$ features $\mathbf{u}_k \in \mathbb{R}^F$ associated with them. The feature matrix $\mathbf{U} \in \mathbb{R}^{n \times F}$ contains all nodes' stacked feature vectors $\mathbf{u}_k$. We define the node neighborhood $N_i = \{v_j | \mathbf{\Tilde{A}}_{ij} > 0\}$ as the set of all nodes connected to $v_i$.

\subsection{Personalized PageRank}
Our approach inherits basic concepts and intuition from personalized PageRank~\citep{page1999pagerank}, which we briefly describe here.
PageRank~\citep{page1999pagerank} was originally introduced to score the importance of webpages for web searches. In their work, webpages represent individual nodes in a graph and links on these webpages are modeled as directed edges between these nodes. The solution to PageRank is the fixed point of the equation 
\begin{equation}
    \label{eq:pr}
    \mathbf{r} = \mathbf{A}\mathbf{r}\, ,
\end{equation}
with $\mathbf{r} \in \mathbb{R}^{n}$ being the dominant eigenvector of $\mathbf{A}$. The vector $\mathbf{r}$ can be obtained by power iteration with an arbitrary initial $\mathbf{r}_0$~\citep{page1999pagerank}. 
For an intuitive interpretation of Eq.~\eqref{eq:pr}, we can interpret $\mathbf{A}$ as the stochastic transition matrix over the graph, providing a connection to a random walk. Therefore the stationary probability distribution induced by a random walk is the same as $\mathbf{r}$ in the limit~\citep{page1999pagerank}.
This also results in $r$ only depending on the graph structure and not on prior information available for nodes. Therefore, the authors also introduce personalized PageRank~\citep{page1999pagerank}
\begin{equation}
    \label{eq:ppr}
    \mathbf{r} = (1-\alpha)\mathbf{A}\mathbf{r} + \alpha\mathbf{u}
\end{equation}
that adds a chance $\alpha$ as a way to teleport back to a personalization vector $\mathbf{u} \in \mathbb{R}^{n}$ representing an initial distribution over all pages. The corresponding interpretation for a random walk is to introduce a chance to reset the random walk to the personalization vector~\citep{page1999pagerank}.

\subsection{Graph Neural Networks}
Another concept we build upon are Graph Neural Networks (GNNs), specifically their subtype of Message-Passing Neural Networks (MPNNs)~\citep{gilmer2017neural}.
GNNs apply permutation equivariant operations to graph structured data in order to identify task-specific features.
Originating from spectral graph convolutions~\citep{hammond2011wavelets} as a localized first-order approximation, each message-passing operation updates the node states $\mathbf{h}_i$ by combining the information of the direct neighborhood $N_i$ for each node $v_i$~\citep{wu2020comprehensive}. 
The general framework can be described as a node-wise update function
\begin{equation}
    \mathbf{h}^{(l+1)}_i = \psi\left(\mathbf{h}^{(l)}_i,\bigoplus_{j \in N_i} \omega(\mathbf{h}^{(l)}_i,\mathbf{h}^{(l)}_j)\right)\, ,
\end{equation}
for each state $\mathbf{h}_i$, using some functions $\omega$ and $\psi$ and a permutation invariant aggregation function $\bigoplus$.
In this work, we will demonstrate our approach using the very basic instantiation of this framework, namely the Graph Convolutional Network (GCN)~\citep{kipf2016semi}. Making use of the normalized adjacency matrix $\mathbf{\Tilde{A}} \in \mathbb{R}^{n \times n}$, the GCN can be expressed in matrix notation
\begin{equation}
\label{eq:gcn}
    \mathbf{H}^{(l+1)} = \phi\left(\mathbf{\Tilde{A}} \mathbf{H}^{(l)} \mathbf{W}^{(l)}\right)
\end{equation}
using a linear transformation $\mathbf{W} \in \mathbb{R}^{d \times d}$ and $\phi$ as an element-wise activation function. $\mathbf{H}^{(l)} \in \mathbb{R}^{n \times d}$ contains the node states $\mathbf{h}^{(l)}_i$ of all nodes after layer $l$. 
Each layer aggregates information only from direct neighborhood $N_i$ for each node $v_i$. Thus, after $k$ such layers, each node only has access to information a maximum of $k$ hops away. 
Given this property, choosing any number $k$ of these layers leads to information at $k+1$ hops away being impossible to be considered for making predictions. Moreover, even when the number of layers $k$ can be selected to be sufficient for all potentially considered graphs, a large number $k$ leads to various additional issues that we will describe next. 

\subsubsection{Over-smoothing.}
\label{sec:oversm}
Recent work found that stacking many layers of Eq.~\eqref{eq:gcn} leads to a degradation of experimental performance that is caused by an effect called over-smoothing~\citep{kipf2016semi,li2018deeper,xu2018representation}. 
~\citet{li2018deeper} show that Eq.~\eqref{eq:gcn} is a special form of Laplacian smoothing leading to node representation becoming more similar the more layers are added. They prove that Laplacian smoothing converges to a linear combination of dominant eigenvectors.
While some smoothing is needed to share information between nodes, representations eventually become indistinguishable with too much smoothing, thus making accurate data-dependant predictions harder~\citep{li2018deeper}.

On a similar note,~\citet{xu2018representation} find a close connection between $k$ layers of Eq.~\eqref{eq:gcn} and a $k$-step random walk. They find that both to converge the limit distribution of the random walk. In the limit, a random walk becomes independent of the root nodes and therefore loses the locality property of individual nodes. Therefore, representations become independent of the starting node and initial node features, thus becoming indistinguishable~\citep{xu2018representation}. 
In practice, the performance of Eq.~\eqref{eq:gcn} already degrades with more than two layers in many cases~\citep{kipf2016semi}.

\subsubsection{Memory Complexity.} 
Another reason that prevents GNNs from being deep models is the memory complexity. Graphs can quickly surpass a million nodes, which leads to out-of-memory issues due to the linear memory requirements $\mathcal{O}(kn)$ in the number of layers $k$ and the number of nodes $n$. 
Several approaches try to lower the memory complexity by only considering samples of nodes from a local neighborhood~\citep{hamilton2017inductive}. Due to an effect known as the neighborhood explosion, the number of nodes in the $k$-hop neighborhood $\mathcal{O}(d^k)$ explodes, with $d$ being the average node degree. 
Thus, for a large number of layers $k$, the benefit vanishes. 
Other approaches cluster the graph into subgraphs and use these for training~\citep{chiang2019cluster,zeng2020graphsaint}, but cannot leverage the full potential of the entire graphs relationships. 
Therefore, this issue needs to be considered when designing deep graph neural networks.
\section{Related Work}
\label{sec:related}
Several approaches aim to increase the depth of MPNNs and simultaneously deal with over-smoothing and memory consumption.
~\citet{rong2020dropedge} found over-smoothing to occur faster for nodes with many incoming edges and propose DropEdge as the equivalent to dropout in regular neural networks. They randomly sample edges to remove during each training epoch and show that the effect of over-smoothing gets slowed down. 
~\citet{klicpera_diffusion_2019} propose a diffusion process that they find to be beneficial for semi-supervised node classification tasks for graphs with high homophily but encounters problems with complex graphs. ~\citet{zhu2020beyond} further discuss the issue of settings with low homophily.
~\citet{li2018deeper} co-train a random walk model that explores the global graph topology.
Inspired by the findings from ResNet~\citep{he2016deep}, .~\citet{chen2020simple} propose GCNII that makes use of residual connections in two ways.
In each layer, they add an initial residual connection to the input state $\mathbf{H}^{(0)}$ and an identity mapping to the weights, which were shown to have beneficial properties~\citep{hardtM17}. 
~\citet{xu2018representation} combine the results of all intermediate iterations in JKNet.
Other works find a rescaling of the weights to alleviate the over-smoothing problem~\citep{Zhao2020PairNorm,oono2020graph}.
While these approaches help reduce the effect of over-smoothing, they are limited in practical depth and the issue still arises.

\subsection{Infinite-Depth Graph Neural Networks}
Evaluating the option of repeating iterations infinitely many times have been analyzed in various approaches~\citep{gori2005new,gallicchio2020fast,klicpera2018predict,gu2020implicit}. These methods iterate some graph convolution until convergence by employing weight-sharing and ensuring the convergence of their formulations.
When using an equation for an infinite-depth GNN, the result needs to converge to a unique solution. We summarize this under the following definition of well-posedness.
\begin{definition} (Well-posedness).
Given an input matrix $\mathbf{X} \in \mathbb{R}^{N \times D}$, an equation $\mathbf{Y} = g(\mathbf{X})$, with $g$ being an infinitely recursive function is well-posed, if
\begin{enumerate}
    \item The solution $\mathbf{Y}$ is unique
    \item $g(\mathbf{X})$ converges to the unique solution $\mathbf{Y}$\, .
\end{enumerate}
\end{definition}

While the GCN (Eq.~\eqref{eq:gcn}) is not generally well-posed, our work proposes a similar equation that we prove to be well-posed.
We start by reviewing two recent approaches to infinitely deep graph neural networks that serve as the starting point for our contribution. The first~\citep{klicpera2018predict} is derivated from the PageRank~\citep{page1999pagerank} algorithm, the other is the fixed-point solution of an equilibrium equation~\citep{gu2020implicit}.

\subsubsection{APPNP.}
~\citet{klicpera2018predict} propose a propagation scheme derived from personalized PageRank~\citep{page1999pagerank}. They identify the connection between the limit distribution of MPNNs and PageRank, with both losing focus on the local neighborhood of the initial state.
 As personalized PageRank was introduced as a solution to this issue for PageRank~\citep{page1999pagerank}, they adopt this idea for MPNNs. 
 They set the personalization vector $\mathbf{r}$ from Eq.~\eqref{eq:ppr} to the hidden state of all nodes $\mathbf{H}^{(0)}$.
 A chance $\alpha$ to teleport back to the root node preserves the local neighborhood with the tunable parameter. ~\citet{klicpera2018predict} transfer this idea to MPNNs with Approximate Personalized Propagation of Neural Predictions (APPNP)~\citep{klicpera2018predict}
 \begin{equation}
\label{eq:appnp}
    \mathbf{H}^{(l+1)} = (1-\alpha)\mathbf{\Tilde{A}} \mathbf{H}^{(l)} + \alpha \mathbf{H}^{(0)}
\end{equation}
that repeatedly, potentially infinitely many times, aggregates the neighborhood. They also add a chance of going back to the initial state $\mathbf{H}^{(0)} = f_{\theta}(\mathbf{U})$, that is be the output of previous layers $f_\theta$. 
They show that Eq.~\eqref{eq:appnp} is well-posed for any $\alpha \in (0,1], \mathbf{H}^{(0)} \in \mathbb{R}^{N \times D}, \mathbf{\Tilde{A}} \in \mathbb{R}^{N \times N}$ with $\mathrm{det}(\mathbf{\Tilde{A}}) \leq 1$. Typical normalizations $\mathbf{\Tilde{A}}$ of the adjacency matrix satisfy this property.
Notably, Eq.~\eqref{eq:appnp} does not utilise any learnable parameters. They rather propose to separate the propagation scheme in Eq.~\eqref{eq:appnp} from the learnable part, by making $\mathbf{H}^{(0)} = f_{\theta}(\mathbf{U})$ as node-wise application of a MLP. This method is proposed only for semi-supervised node classification tasks, with a softmax activation employed to transform the output of the last iteration $\mathbf{H}^{(K)}$ of Eq.~\eqref{eq:appnp} to class predictions.

\subsubsection{Implicit Graph Neural Networks.}
Independently, ~\citet{gu2020implicit} propose the Implicit Graph Neural Network (IGNN) by adapting the general implicit framework~\citep{el2021implicit} for graph convolutions. 
They obtain the fixed-point solution of a non-linear equilibrium equation
\begin{equation}
\label{eq:ignn}
    \mathbf{X} = \phi(\mathbf{W}\mathbf{X}\mathbf{\Tilde{A}} + f_{\theta}(\mathbf{U}))
\end{equation}
by iterating it until convergence. While not being well-posed in general, they prove the well-posedness of Eq.~\eqref{eq:ignn} for the specific case that $\lambda_{pf}(|\mathbf{A}^T \otimes \mathbf{W}|) < 1$ with $\lambda_{pf}$ being the Perron-Frobenius (PF) eigenvalue. They make use of the Kronecker product $\otimes$ and the Perron-Frobenius theory~\citep{berman1994nonnegative}. 
Since $\mathbf{\Tilde{A}}$ is fixed, the matrix of parameters $\mathbf{W}$ needs to be strictly constrained to fulfill $\lambda_{pf}(|\mathbf{A}^T \otimes \mathbf{W}|) < 1$.
The set $\mathcal{M}$ of allowed matrices $\mathbf{W}$ forms an $\mathcal{L}_1$-ball, with any weight matrix outside the ball not leading to convergence.
Remaining inside this ball cannot be guaranteed by regular gradient descent. Instead, after each step of regular gradient descent, they project the result to the closest point on the ball using projected gradient descent, for which efficient algorithms exist~\citep{duchi2008efficient}.
While their inspiring work shows great experimental results, we identify a couple of shortcomings with.
Many weight matrices cannot be used given the strict constraint on $\mathbf{W}$, hindering the model capacity. Further, the projection onto the $\mathcal{L}_1$-ball changes the direction of the gradient update away from the steepest descent. Therefore optimization steps are less effective in reducing the models' loss. The strict constraint and the resulting projection step also add complexity to the method's theoretical derivation and practical implementation. Considering different neighborhood sizes was found to be important when applying graph algorithms to varying graph types~\citep{abu2018watch,grover2016node2vec}, not having a way to control the effective depth of the model is also unsatisfying. 
\section{PageRank Graph Neural Network}
\label{sec:methods}
The solution of PageRank is the stationary probability distribution that is independent of the input. Given the close relation between PageRank (Eq.~\eqref{eq:pr}) and MPNNs (Eq.~\eqref{eq:gcn}), the locality of the data and the influence of the input features also diminish with a MPNN, as identified by~\citep{klicpera2018predict}.
As personalized PageRank was introduced to prevent the loss of focus for PageRank~\citep{page1999pagerank}, we introduce the Personalized PageRank Graph Neural Network (PPRGNN) based on personalized PageRank, that similarly assures the locality of the node states in the limit. 
Using the initial state $f_\theta(\mathbf{U})$ as personalization matrix for teleportation~\citep{klicpera2018predict}, PPRGNN can be understood as repeatedly applying graph convolutions with a chance to teleport back to this initial state. 
We assure the convergence of PPRGNN to a unique solution, so our method allows an arbitrary amount of layers - potentially infinitely many - without suffering from over-smoothing. 
Practically, we iterate graph convolutions until further iterations have negligible impact and our solution is close to the limit distribution.
In this work, we adopt GCNs~\citep{kipf2016semi}, which are the basic version of MPNNs, but these are directly replaceable by more advanced types.

We denote the chance of traversing the graph further by $\alpha_l$. 
Rewriting the formulation of the GCN in a similar way to personalized PageRank, we come up with our formulation
\begin{equation}
\label{eq:invert}
    \mathbf{H}^{(l+1)} = \phi\left(\alpha_{l}\mathbf{\Tilde{A}}\mathbf{H}^{(l)}\mathbf{W} + f_\theta(\mathbf{U})\right)
\end{equation}
with $\mathbf{H}^{(0)} = \mathbf{0}$ that utilizes shared and unconstrained parameters $\mathbf{W}$.
Due to the recursive nature and no constraints, exponential growth in $\mathbf{W}$ prevents well-posedness for any fixed $\alpha_{l}$.
The issue with having no guarantees for convergence is that the furthest distant nodes are multiplied with the highest exponential of $\mathbf{W}$, which potentially dominates the result. As in PageRank, these only depend on the graph structure and not on the node features, leading to the loss of locality of the resulting node features.

Our core idea to guarantee convergence of Eq.~\eqref{eq:invert} without constraining the parameters as in~\citet{gu2020implicit} is to reduce the chance of expanding further $\alpha_{l}$ with the distance to the root node. As the message-passing formulation is connected to a random walk on the graph, another interpretation is to increase the chance of resetting the random walk with the number of steps taken. When $n$ is the number of steps taken in that walk, we find using a decay of $\alpha_n=\frac{1}{n}$ to be sufficient for converging to a unique solution.
The recursive nature of our formulation leads to a multiplication of all $\alpha_n$, resulting in the influence to decay by $\frac{1}{n!}$. Because the recursive application of $\mathbf{W}$ only leads to an exponential $\mathbf{W}^n$ growth, the equation converges.
For control over the effective depth, i.e., the speed of convergence and numerical stability, we use $\frac{1}{1+n\epsilon}$ and formally prove its convergence for any $\epsilon > 0$ later. 
We set the value for teleporting back to $f_\theta(\mathbf{U})$ fixed to $1$ because in the limit the chance $(1-\alpha_l)$ would become very small for close neighbors, leading to the same issues of over-smoothing that we described in section~\ref{sec:oversm}.

Setting $\alpha_{l}$ in Eq.~\eqref{eq:invert} accordingly to our findings, the following issue arises:
The most distant nodes are processed first in Eq.~\eqref{eq:invert}, and the direct neighbors are processed in the last iteration. 
This results from recursively applying the adjacency matrix $\mathbf{\Tilde{A}}$ on the input, leading to the initial state being transformed $k$ times by $\mathbf{\Tilde{A}}$.
Thus, for calculating $\mathbf{H}^{(1)}$, the expansion factor $\alpha_0$ needs to be minimal, which is the opposite of using the iteration $l$ as $n$.

In case we are interested in a fixed number $k$ of total iterations, we can directly set $\alpha_{l}=\frac{1}{1+(k-l-1)\epsilon}$ for each layer $l$. 
When using a fixed number of iterations, this approach is ready for usage directly. Since we are interested in the case when $k \to \infty$, starting with $\alpha_0$ poses a challenge.

For a theoretical analysis of the convergence of Eq.~\eqref{eq:invert}, an equation that can be iterated infinitely-deep independently of $k$ is desired. We achieve this by setting the index variable to $n=k-l$ resulting in the flipped equation
\begin{equation}
\label{eq:ours}
    \mathbf{G}^{(n)} = \phi\left(\beta_n\mathbf{\Tilde{A}}\mathbf{G}^{(n+1)}\mathbf{W} + f_\theta\left(\mathbf{U}\right)\right)
\end{equation}
with $\beta_n = \alpha_{k-l-1}$ that is semantically unchanged, i.e., $\mathbf{G}^{(0)} = \mathbf{H}^{(k)}$ for any $k$ used for both $\mathbf{G}$ and $\mathbf{H}$. Calculating $\mathbf{G}^{(n)}$ from a given $\mathbf{G}^{(n+1)}$ can be performed without knowing $k$ beforehand, helping us in the theoretical analysis by expanding the recursive equation infinitely deep without the need to set a fixed value for $k$.
It also leads to a cleaner proof of convergence, which we will provide next.
We further simplify our notation by denoting $\mathbf{G}^{(l;k)}$ as the result of $k$ iterations performed by setting $\mathbf{G}^{k+l+1}= \mathbf{0}$, resulting in $\mathbf{G}^{(l)}$.

\begin{theorem}
\label{the:main}
The result of $\mathbf{G}^{(l;k)}$ using the equation $\mathbf{G}^{(n)} = \phi\left(\beta_n\mathbf{\Tilde{A}}\mathbf{G}^{(n+1)}\mathbf{W} + \mathbf{B}\right)$ with $\beta_n=\frac{1}{1+n\epsilon}$ converges to a unique solution when $k \to \infty$ for any $l \in \mathbb{R}$ $\mathbf{W} \in \mathbb{R}^{d \times d}, \mathbf{\Tilde{A}} \in \mathbb{R}^{n \times n}, \mathbf{B} \in \mathbb{R}^{d \times n}, n \in \mathbb{N}, \epsilon > 0$, any Lipschitz continuous activation function $\phi$.
\end{theorem}

We refer to the supplementary material for all proofs.

Practically, for either Eq.~\eqref{eq:invert} or Eq.~\eqref{eq:ours} processing starts at the furthest distant nodes, for which $k$ needs to be known.
This a challenge, because we do not know beforehand when our convergence criterion is satisfied.
As our interest is in the limit state $\mathbf{G}^{(0;k)}$ when $k \to \infty$, we make use of a convergence threshold $\epsilon$ to identify the number of required iterations 
\begin{equation}
    k = \min \{M \mid \mathbf{G}^{(0;k)} - \mathbf{G}^{(0;k+1)} < \epsilon\}
\end{equation}
until our solution is close to the limit and iterating further has negligible impact. 
Because even the initial iteration $\mathbf{G}^{(k-1;1)} \neq \mathbf{G}^{(k;1)}$ is different, intermediate results from $\mathbf{G}^{(0;k-1)}$ cannot be reused for computing $\mathbf{G}^{(0;k)}$. A full recalculation is needed, which requires $k!$ iterations. 

Instead, we take a different route to determine $k$.
Determining at which iteration the difference of expanding further on the graph becomes negligible is approximately the same as determining how far the influence of nodes in the graph reach using our message passing scheme.
To determining this, we ignore the teleportation term and estimate the influence of the initial state $f_\theta(\mathbf{U})$ on the result of $l$ iterations $\mathbf{G}^{(0;l)}$ with the equation
\begin{equation}
\label{eq:conver}
    \mathbf{E}^{(l+1)} = \phi(\alpha_{l+1}\mathbf{\Tilde{A}}\mathbf{E}^{(l)}\mathbf{W})
\end{equation}
by setting $\mathbf{E}^{(0)} = f_\theta(\mathbf{U})$.
Unlike in Eq.~\eqref{eq:invert} where we reversed the equation, the result $\mathbf{E}^{(m)}$ is equal for $\alpha_l=\frac{1}{1+(m-l-1)\epsilon}$ and $\alpha_l=\frac{1}{1+l\epsilon}$ when we use ReLU as $\phi$. Note, that we start with $\alpha_{l+1}$ because this is the first $\alpha$ that is applied to the teleportation matrix $f_{\theta}(\mathbf{U})$. Eq.~\eqref{eq:conver} converges for similar reasons as Eq.~\eqref{eq:ours}, only towards $\mathbf{0} \in 0^{d \times n}$, which we proof with the following theorem.
\begin{theorem}
\label{the:forward}
The equation $\mathbf{E}^{(l+1)} = \phi\left(\alpha_l\mathbf{\Tilde{A}}\mathbf{E}^{(l)}\mathbf{W}\right)$ with $\alpha_l=\frac{1}{1+l\epsilon}$ converges to $\mathbf{0} \in 0^{d \times n}$ for any $\mathbf{W} \in \mathbb{R}^{d \times d}, \mathbf{\Tilde{A}} \in \mathbb{R}^{n \times n}, l \in \mathbb{N}, \epsilon > 0$, any initial $\mathbf{E}^{(0)}$ and the ReLU activation function $\phi$. The solution can be obtained by iterating the equation. For any fixed number of iterations $m$, the solution $\mathbf{E}^{(m)}$ is the same as using $\alpha_l=\frac{1}{1+(m-l-1)\epsilon}$.
\end{theorem}

Since we can evaluate Eq.~\eqref{eq:conver} directly by iterating until our convergence criterion is met, we find the required number of steps with 
\begin{equation}
k^{\prime} = \min\{l \mid \mathbf{E}^{(l)} < \epsilon\}\, .
\end{equation}
At this point the effect of the initial state on nodes of distance $l$ is negligible. 
We use $k^{\prime}$ as $k$ in Eq.~\eqref{eq:ours} and execute the forward pass. The result $\mathbf{H}^{(k^{\prime})}$ gets passed onto the next operation in our model, as with other graph convolutions.

\subsection{Efficient Optimization}
While we do not use Eq.~\eqref{eq:conver} for gradient computation, even tracking only Eq.~\eqref{eq:invert} with autograd software would still lead to memory consumption that is linear in the number of layers.
Similarly as in the forward pass, the gradients converge to $\mathbf{0}$ for distant nodes. We iterate the computation of gradients until the same convergence criterion is met.
Because of faster converge in the backward pass, this allows the optimization of the model with reduced memory consumption. We will further limit the iterations to guarantee constant complexity, independently of the number of iterations performed. 

For calculating derivatives we use the reformulation from Eq.~\eqref{eq:ours}.
We introduce additional notation and set $\mathbf{\hat{Y}} = f_\theta(\mathbf{G}^{(0)})$ as the output of our model, $\mathbf{Y}$ as the target, and $\mathcal{L} = l(\mathbf{\hat{Y}}, \mathbf{Y})$ to be our loss calculated with any differentiable loss function $l$. We are interested in the partial derivatives of our loss $\mathcal{L}$ with respect to the parameters $\mathbf{W}$ and the input state $\mathbf{B}$. We let autograd solve the derivation $\frac{\partial L}{\partial \mathbf{G}^{(0)}}$ and apply the chain rule for other partial derivatives. To simplify our notation for the application of the chain rule, we further define $\mathbf{Z}^{(n)}=\alpha_n \mathbf{\Tilde{A}}\mathbf{G}^{(n+1)}\mathbf{W} + \mathbf{B}$ and $\mathbf{G}^{(n)} = \phi(\mathbf{Z}^{(n)})$. All further partial derivatives can be calculated by using the following equations:

\begin{equation}
    \frac{\partial L}{\partial \mathbf{G}^{(n)}} = \alpha_n \mathbf{\Tilde{A}}^T\frac{\partial L}{\partial \mathbf{Z}^{(n-1)}}\mathbf{W}^T
\end{equation}

\begin{equation}
    \frac{\partial L}{\partial \mathbf{Z}^{(n)}} = \phi^{\prime}\left(\alpha_n \mathbf{\Tilde{A}}\mathbf{G}^{(n+1)}\mathbf{W} + \mathbf{B}\right) \odot \frac{\partial L}{\partial \mathbf{G}^{(n)}}
\end{equation}

\begin{equation}
    \frac{\partial L}{\partial \mathbf{W}} = \sum_{n=0}^{\infty} \alpha_n \left(\mathbf{\Tilde{A}}\mathbf{G}^{(n+1)}\right)^T\frac{\partial L}{\partial \mathbf{Z}^{(n)}}
    \label{eq:dW}
\end{equation}

\begin{equation}
    \frac{\partial L}{\partial \mathbf{B}} = \sum_{n=0}^{\infty} \frac{\partial L}{\partial \mathbf{Z}^{(n)}}
    \label{eq:dB}
\end{equation}
The partial derivatives $\frac{\partial L}{\partial \mathbf{W}}$ and $\frac{\partial L}{\partial \mathbf{B}}$ converge for similar reasons as before, so we iterate Eq.~\eqref{eq:dW} and Eq.~\eqref{eq:dB} until our convergence criterion is met.
The convergence rate turns out to be much faster than the convergence rate of the forward pass, which results in reduced practical memory consumption.
To theoretically guarantee constant memory consumption, we only consider a fixed amount $n$ of elements in the sum, similarly to the effectiveness of Truncated Backpropagation Through Time (TBPTT)~\citep{sutskever2013training} for sequential data. This also reduces the time complexity of the backward pass to be constant.
We found this restriction to have negligible impact even for small values of $N$.
Depending on available memory, we either store the intermediate results for gradient computation or use gradient checkpointing~\citep{chen2016training} with a few additional forward iterations.
Note, that for the backward step the solutions of $\mathbf{G}^{(0)},\dots,\mathbf{G}^{(n)}$ are needed explicitly. We assure the convergence of all used $\mathbf{G}^{(i)}$ by using the fact that $\mathbf{G}^{(n;k)} < \mathbf{G}^{(0;k)}$ and therefore compute $\mathbf{G}^{(0;k+n)}$ instead of $\mathbf{G}^{(0;k)}$ in the initial forward pass.
\newfloatcommand{capbtabbox}{table}[][\FBwidth]
\section{Experiments}
\label{sec:experiments}
We evaluate the effectiveness of PPRGNN on various public benchmark datasets and compare these results to popular methods and other infinite-depth approaches. 
We evaluate our approach on an inductive node classification task, a transductive node classification task, and five graph classification tasks. Table~\ref{tab:properties} shows detailed properties of all used datasets.
We closely follow the experimental settings of IGNN~\citep{gu2020implicit} and inherit their architectures, only replacing their formulation directly with ours. Thus the number of parameters is the same, so the comparison with their approach is the most meaningful for us. We apply APPNP to all tasks using their setup with $10$ iterations. We further compare PPRGNN with a series of other popular methods for the tasks of node classification and graph classification and reuse the results reported in~\citet{gu2020implicit}.
Due to the increased modeling capacity, we use gradient clipping and weight decay. Additionally, we tune $\epsilon$, the learning rate and whether self-loops are taken into account for $\mathbf{\Tilde{A}}$ for the three different tasks. We set $n=5$ for the backward pass. We reduce the learning rate when the training loss plateaus. All experiments are executed on a single Nvidia Tesla P100.

\begin{table}[tb]
  \floatsetup{floatrowsep=qquad}
  \begin{floatrow}[2]
    
    \ttabbox%
    {\begin{tabular}{c|c|c|c}
\toprule
    Dataset & \# of Graphs & Avg. \# of nodes & \# of classes \\
    \midrule
    Amazon & 1 & \num{334863} & 58 \\
    PPI & 22 & 2373 & 121 \\
    MUTAG & 188 & 17.9 & 2 \\
    PTC & 344 & 25.5 & 2 \\
    COX2 & 467 & 41.2 & 2 \\
    PROTEINS & 1113 & 39.1 & 2 \\
    NCI1 & 4110 & 29.8 & 2 \\
    \bottomrule
\end{tabular}}
    {\caption{Properties of datasets used for evaluation.}\label{tab:properties}}
    \hfill%
    \ttabbox%
    {\begin{tabular}{c|c}
\toprule
     Method & Micro-$F_1$-Score \\
     \midrule
     MLP & 46.2 \\
     GCN & 59.2 \\
     SSE & 83.6 \\
     GAT & 97.3 \\
     IGNN & 97.6 \\
     APPNP & 44.8 \\
     \midrule
     PPRGNN & \textbf{98.9} \\
     \bottomrule
\end{tabular}}
    {\caption{Micro-$F_1$-Scores for PPI.}\label{tab:ppi}}
  \end{floatrow}
\end{table}%

\subsubsection{PPI}
We consider the task of role prediction of proteins in graphs of protein-protein interactions (PPI)~\citep{hamilton2017inductive}.
In this inductive node classification task, we use $18$ graphs for training our model, $2$ for validation, and $2$ for testing. Our data split matches that in previous work~\citep{hamilton2017inductive}. As taken over from IGNN, our model consists of $5$ stacked layers, each iterating until convergence. We set $\epsilon=0.25$ and find self-loops detrimental to our approach.
In addition to IGNN and APPNP, reference methods are a MLP, GCN~\citep{kipf2016semi}, SSE~\citep{dai2018learning}, GAT~\citep{velickovic2018graph}.
The Micro-$F_1$-Scores for all considered approaches are presented in Table~\ref{tab:ppi}.
PPRGNN outperforms all of these approaches and reduces the error by more than $50\%$ compared to IGNN.
Our trained PPRGNN uses a total of $82$ message passing iterations in testing, while GCN and GAT use a maximum of $3$ iterations.
We also compare the time needed for PPRGNN to surpass the Micro-$F_1$-Score of IGNN in Figure~\ref{tab:time}. PPRGNN needs fewer iterations and also takes less time per Epoch. This comes from accurate gradient descent steps without projection and being able to adjust the speed of convergence with $\epsilon$.
We find APPNP to underfit the data due to the limited modeling capacity, even when the number of parameters uses all available memory. 

\begin{table}[tb]
\centering
\caption{Time and epochs needed until PPRGNN surpasses the best epoch of IGNN on the validation set.}
\begin{tabular}{c|c|c|c|c}
\toprule
     Dataset & Method & Epochs & Avg. Time per Epoch & Total Time \\
     \midrule
     \multirow{2}{*}{Amazon (0.05)} & IGNN & 872 & 14s & 3h 21m  \\
     & PPRGNN & \textbf{175} & \textbf{11s} & \textbf{32m} \\
     \midrule
     \multirow{2}{*}{PPI} & IGNN & 58 & 26s & 25m \\
     & PPRGNN & \textbf{47} & \textbf{18s}  & \textbf{14m} \\
     \bottomrule
\end{tabular}
\label{tab:time}
\end{table}

\subsubsection{Amazon}
\begin{figure}[tb]
  \begin{subfigure}[b]{0.4\textwidth}
    \includegraphics[width=\textwidth]{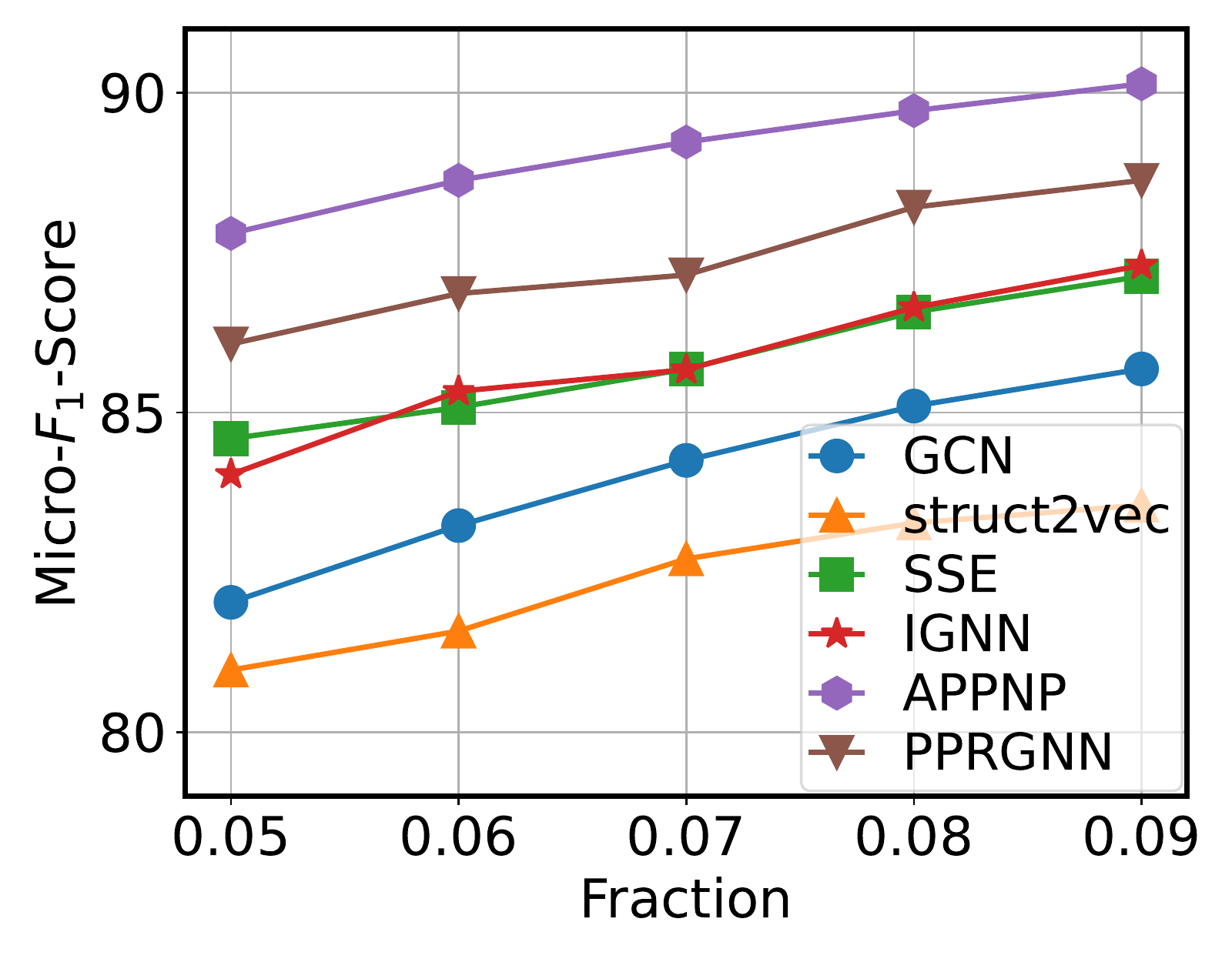}
    \caption{Micro-$F_1$-Scores.}
    \label{fig:f1}
  \end{subfigure}
  \hfill
  \begin{subfigure}[b]{0.4\textwidth}
    \includegraphics[width=\textwidth]{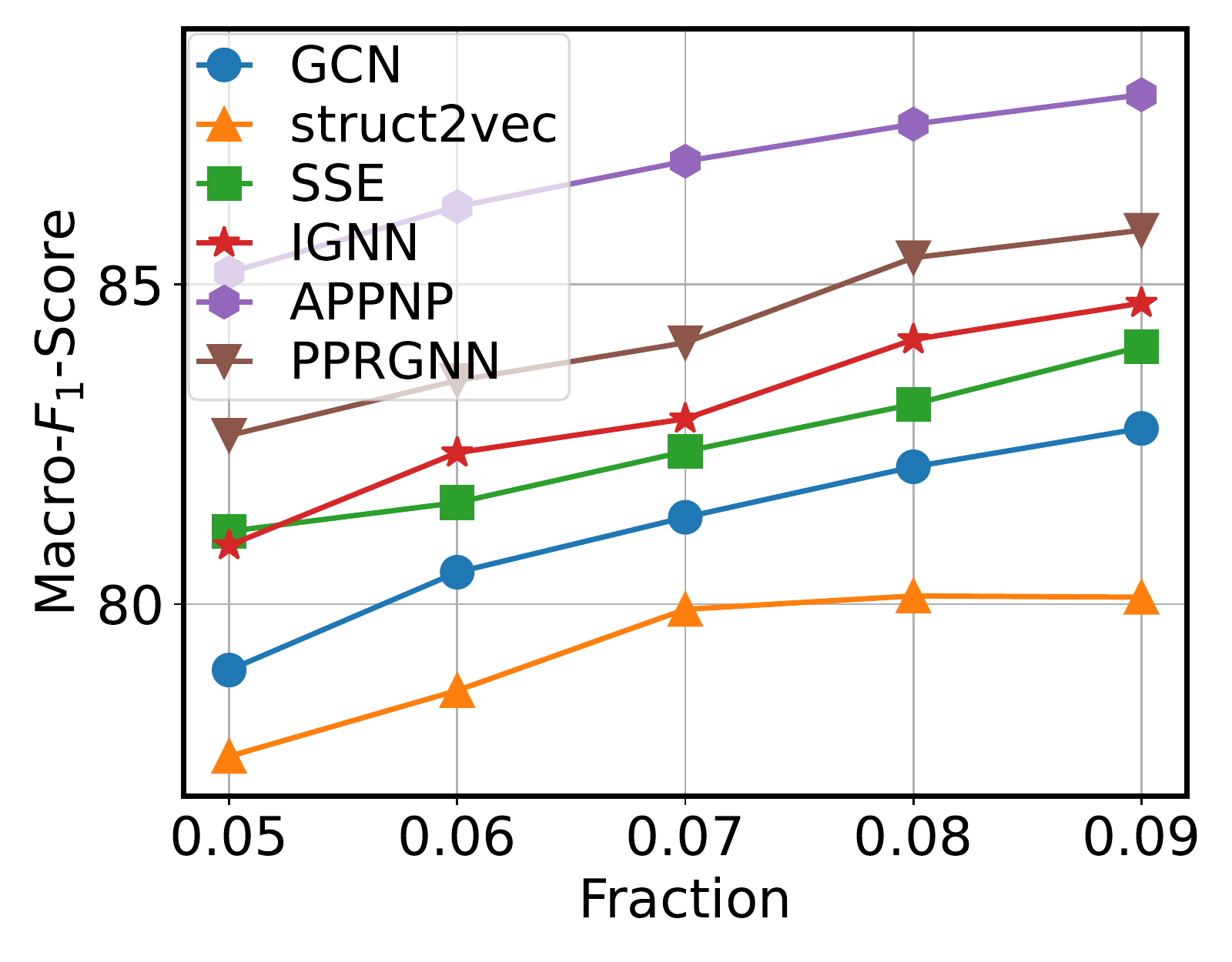}
    \caption{Macro-$F_1$-Scores.}
    \label{fig:f2}
  \end{subfigure}
  \caption{Comparison of results on the Amazon dataset. The fraction of labels used for optimization varies between $0.05$ and $0.09$.}
  \label{fig:amazon}
\end{figure}

To test the scalability of our approach, we apply it to the Amazon product co-purchasing network data set~\citep{yang2015defining}. 
Following the settings from~\citet{dai2018learning}, product types with at least \num{5000} different products are selected.
This results in \num{334863} nodes representing products and \num{925872} edges representing products that have been purchased together. The task is to predict the correct product type for each node. 
Nodes do not have any features, so predictions are made solely based on the graph structure.
We use the same data split as~\citet{dai2018learning}, leading to a fraction of nodes used for training varying between $5\%$ and $9\%$. A fixed set consisting of $10\%$ of the nodes is used for training, the rest for validation.
The main challenge of this task is not the prediction complexity but rather dealing with the sparsity of the available labels.
Our architecture consists of our PPRGNN layer combined with a linear operation before and after.
We compare our results with APPNP and reuse the result found in~\citet{dai2018learning,gu2020implicit} for IGNN~\citep{gu2020implicit}, SSE~\citep{dai2018learning}, struct2vec~\citep{ribeiro2017struc2vec} and GCN~\citep{kipf2016semi}.
Micro-$F_1$-Scores and Macro-$F_1$-Scores are shown in Figure~\ref{fig:amazon} for varying fractions of labels used. While we outperform IGNN, SSE, struct2vec and GCN across all settings by at least $1\%$, APPNP performs the best. We find the low modeling capacity of APPNP to be better suited for generalizing in this scenario. Again, we compare the execution time needed for PPRGNN to outperform IGNN (Figure~\ref{tab:time}) and find PPRGNN to converge in fewer epochs, with each epoch executing faster. This further adds to our point of benefiting from applying gradient descent without projection and controlling convergence speed.

\subsubsection{Graph Classification}
We now evaluate our approach for the task of graph classification on five open graph datasets, namely MUTAG, PTC, COX2, PROTEINS, and NCI1. Following the same setup from previous work, we conduct a 10-fold cross-validation for each dataset and report the mean and standard deviation of the folds validation sets. 
We integrate our formulation with $\epsilon=1$ into the architecture from IGNN, consisting of $3$ stacked iterations until convergence.
For regularization, we add a weight decay of $1\mathrm{e-}6$ and gradient clipping of $25$ to all datasets. For NCI1, we find removing self-loops to be helpful for generalization.
For comparison, we use several graph kernel approaches (GK~\citep{shervashidze2009efficient}, RW~\citep{gartner2003graph}, WL~\citep{shervashidze2011weisfeiler}) and GNN approaches (DGCNN~\citep{zhang2018end}, GCN~\citep{kipf2016semi}, GIN~\citep{xu2018powerful}) in addition to IGNN and APPNP. We reuse reported results from~\citet{gu2020implicit}.
PPRGNN outperforms all other approaches across $4$ out of $5$ datasets by at least $1\%$ and is the second-best performing model with competitive accuracy on the fifth dataset.
Despite using the same $\epsilon=1$ across all experiments, the effective depth ranges from $22$ to $41$ for different datasets. Depth is adaptive even within individual datasets, depending on learned parameters, the examined graph and present node features.
These results further demonstrate the effectiveness of our approach and the potential to create deeper models on a wide variety of datasets.
\begin{table}[tb]
\caption{Comparison of accuracies on various graph classification tasks.}
\centering
\begin{tabular}{c|c|c|c|c|c}
\toprule
     Dataset & MUTAG & PTC & COX2 & PROTEINS & NCI1  \\
     \midrule
     GK & 81.4 \textpm 1.7 & 55.7 \textpm 0.5 & - & 71.4 \textpm 0.3 & 62.5 \textpm 0.3 \\
     RW & 79.2 \textpm 2.1 & 55.9 \textpm 0.3 & - & 59.6 \textpm 0.1 & - \\
     WL & 84.1 \textpm 1.9 & 58.0 \textpm 2.5 & 83.2 \textpm 0.2 & 74.7 \textpm 0.5 & \textbf{84.5} \textpm 0.5 \\
     DGCNN & 85.8 & 58.6 & - & 75.5 & 74.4 \\
     GCN & 85.6 \textpm 5.8 & 64.2 \textpm 4.3 & - & 76.0 \textpm 3.2 & 80.2 \textpm 2.0 \\
     GIN & 89.0 \textpm 6.0 & 63.7 \textpm 8.2 & - & 75.9 \textpm 3.8 & 82.7 \textpm 1.6 \\
     IGNN & 89.3 \textpm 6.7 & 70.1 \textpm 5.6 & 86.9 \textpm 4.0 & 77.7 \textpm 3.4 & 80.5 \textpm 1.9 \\
     APPNP & 87.7 \textpm 8.6 & 64.5 \textpm 5.1 & 82.2 \textpm 5.5 & 78.7 \textpm 4.8 & 65.9 \textpm 2.7 \\
     \midrule
     PPRGNN & \textbf{90.4} \textpm 7.2 & \textbf{75.0} \textpm 5.7 & \textbf{89.1} \textpm 3.9 & \textbf{80.2} \textpm 3.2 & 83.5 \textpm 1.5 \\
     \bottomrule
\end{tabular}
\label{tab:gc}

\end{table}
\section{Conclusion}
\label{sec:conclusion}
We introduced PPRGNN, a reformulation of MPNNs based on personalized PageRank that assures localization and prevents over-smoothing of node features even when using infinitely many layers.
Theoretically based on the personalized version of PageRank which allows teleporting back to the initial state, we adopt this idea for MPNNs, specifically for the basic type GCNs~\citep{kipf2016semi}. Starting from the classic algorithm, we follow intuitive steps to introduce learnable parameters and still converge to a limit distribution.
Compared to previous infinite-depth GNNs, our approach has a higher modeling capacity as we do not place any constraints.
Our empirical evaluation on tasks for graph classification, and inductive and transductive node classification 
confirm our theoretical base. Against regular GCNs that have no way to teleport back to the initial state, we find large improvements across all datasets. We even outperform other comparable approaches, including previous infinite-depth models, across almost all datasets by decent margins.
Despite the theoretical infinite-depth, we introduced a path for efficient optimization, running linearly in the number of layers and only using constant memory. Our formulation allows controlling the convergence rate, leading to considerable improvements in experimental execution time compared to IGNN, a previous infinite-depth model. 
While we show that our formulation allows infinitely many layers, even fixed sized models should benefit from adopting our idea. Our approach is directly applicable to other types of MPNNs, for which our proofs of convergence should hold.

\bibliographystyle{unsrtnat}
\bibliography{bibtex}  
\clearpage
\section{Appendix}
\label{sec:appendix}
\subsection{Proof of Theorem~\ref{the:main}}
\label{sec:proofmain}
\begin{proof}
We show that the solution to Eq.~\eqref{eq:ours} is unique and iterating the equation converges to that solution. As a simplification, we use vector notation $\mathbf{x} = vec(\mathbf{X})$, where the columns of a matrix $\mathbf{X} \in \mathbb{R}^{m \times n}$ are stacked column by column to form a single vector $x \in \mathbb{R}^{mn}$. Equivalently to our matrix notation $\mathbf{G}^{(l;K)}$, $\mathbf{g}^{(l;K)}$ describes the result $\mathbf{g}^{(l)}$ after performing $k$ iterations. Performing $0$ iterations $\mathbf{g}^{(l;0)}=\mathbf{b}$ corresponds to the initial state.
We rewrite Eq.~\eqref{eq:ours} with vector notation to be
\begin{equation}
    \mathbf{g}^{(n)} = \phi(\beta_n \mathbf{M} \mathbf{g}^{({n+1})} + \mathbf{b})
\end{equation}
where $\mathbf{g}^{(n)} = vec(\mathbf{G}^{(n)}), \mathbf{g}^{(n+1)} = vec(\mathbf{G}^{(n+1)}), \mathbf{b} = vec(\mathbf{B})$ and by setting $\mathbf{M} = \mathbf{W}^T \otimes \mathbf{\Tilde{A}}$ following the properties of the kronecker product.
First, we show the uniqueness of the result by proving that the difference
\begin{equation}
\label{eq:unique}
    \lim_{n\to\infty} |\mathbf{g}^{(0)}_p - \mathbf{g}^{(0)}_q| = \mathbf{0}
\end{equation}
between two obtained solutions converges to zero. Next, we show that the difference
\begin{equation}
    \lim_{n\to\infty} |\mathbf{g}^{(0;n)} - \mathbf{g}^{(0;n+1)}| = \mathbf{0}    
\end{equation}
between $n$ iteration steps and further iteration steps converges to zero. We first show how difference of iteration $k$ compares to iteration $k+1$: 
\begin{equation}
\begin{split}
    |\mathbf{g}_p^{(k)} - \mathbf{g}_p^{(k)}| & = |\phi(\beta_k \mathbf{M} \mathbf{g}_p^{(k+1)} + \mathbf{b}) - \phi(\beta_k \mathbf{M} \mathbf{g}_q^{(k+1)} + \mathbf{b})| \\
    & \leq |\beta_k \mathbf{M} \mathbf{g}_p^{(k+1)} + \mathbf{b} - (\beta_k \mathbf{M}\mathbf{g}_q^{(k+1)} + \mathbf{b})| \\
    & \leq |\beta_k\mathbf{M}\mathbf{g}_p^{(k+1)} - \beta_k\mathbf{M}\mathbf{g}_q^{(k+1)}| \\
    & \leq |\beta_k\mathbf{M}||\mathbf{g}_p^{(k+1)} - \mathbf{g}_q^{(k+1)}| \\
    & = \beta_k|\mathbf{M}||\mathbf{g}_p^{(k+1)} - \mathbf{g}_q^{(k+1)}|
\end{split}
\end{equation}
Now we show the uniqueness of an obtained solution by proving Eq.~\eqref{eq:unique} for $\mathbf{g}^{(l)}_p$ and $\mathbf{g}^{(l)}_q$:
\begin{equation}
\begin{split}
      \mathbf{0} & \leq |\mathbf{g}^{(l)}_p - \mathbf{g}^{(l)}_q| \leq \beta_l|\mathbf{M}||\mathbf{g}^{(l+1)}_p - \mathbf{g}^{(l+1)}_q| \\  
        & \leq \beta_l|\mathbf{M}|\beta_{l+1}|\mathbf{M}||\mathbf{g}^{(l+2)}_p - \mathbf{g}^{(l+2)}_q| \\
        & \leq \beta_l\beta_{l+1}|\mathbf{M}|^2|\mathbf{g}^{(l+2)}_p - \mathbf{g}^{(l+2)}_q| \\
    & \leq \dots \\
    & \leq \beta_l\beta_{l+1}\dots\beta_{l+m}|\mathbf{M}|^{m+1}|\mathbf{g}^{(l+m+1)}_p - \mathbf{g}^{(l+m+1)}_q| \\
    & \leq \frac{|\mathbf{M}|^{m+1}}{\frac{(l+m)!}{(l-1)!}\epsilon^{m+1}}|\mathbf{g}^{(l+m+1)}_p - \mathbf{g}^{(l+m+1)}_q| \\
    & \leq \frac{|\mathbf{M}|^{m+1}}{(m+1)!\epsilon^{m+1}}|\mathbf{g}^{(l+m+1)}_p - \mathbf{g}^{(l+m+1)}_q| \\
    & \lim_{m\to\infty}\frac{|\mathbf{M}|^{m+1}}{(m+1)!\epsilon^{m+1}}|\mathbf{g}^{(l+m+1)}_p - \mathbf{g}^{(l+m+1)}_q| = \mathbf{0}
    \end{split}
\end{equation} 
The prove for convergence of Eq.~\eqref{eq:ours} is very similar:
\begin{equation}
\begin{split}
      \mathbf{0} & \leq |\mathbf{g}^{(l;m)} - \mathbf{g}^{(l;m+1)}| \leq \beta_l|\mathbf{M}||\mathbf{g}^{(l+1;m-1)}_p - \mathbf{g}^{(l+1;m)}_q| \\  
    & \leq \frac{|\mathbf{M}|^{m+1}}{(m+1)!\epsilon^{m+1}}|\mathbf{g}^{(l+m+1;0)}_p - \mathbf{g}^{(l+m+1;1)}_q| \\
    & = \frac{|\mathbf{M}|^{m+1}}{(m+1)!\epsilon^{m+1}}|\mathbf{b} - \beta_{l+m+1}\mathbf{M}\mathbf{b} + \mathbf{b}| \\
    & \lim_{m\to\infty}\frac{|\mathbf{M}|^{m+1}}{(m+1)!\epsilon^{m+1}}|\mathbf{b} - \beta_{l+m+1}\mathbf{M}\mathbf{b} + \mathbf{b}| = \mathbf{0} 
    \end{split}
\end{equation}
\hfill \qed
\end{proof} 

\subsection{Proof of Theorem~\ref{the:forward}}
\begin{proof}
The proof for convergence of Eq.~\eqref{eq:ours} is almost also vaild for the convergence of Eq.~\eqref{eq:conver}. The main difference being the convergence to $\mathbf{0}$ instead of an unknown unique point. We use the same notation as before.
We rewrite Eq.~\eqref{eq:conver} with vector notation to be
\begin{equation}
    \mathbf{e}_n = \phi(\alpha_n \mathbf{M} \mathbf{e}_{n+1})
\end{equation}
where $\mathbf{e}^{(n)} = vec(\mathbf{E}^{(n)}), \mathbf{e}^{(n+1)} = vec(\mathbf{E}^{(n+1)})$ and by setting $\mathbf{M} = \mathbf{W}^T \otimes \mathbf{\Tilde{A}}$ following the properties of the kronecker product.
The key to this proof is the inequation
\begin{equation}
    |\phi(x)| \leq |x|
\end{equation}
valid for the ReLU activation function $\phi$.
This leads to the connection 
\begin{equation}
    |\mathbf{e}^{(k)}| \leq \alpha_k|\mathbf{M}||\mathbf{e}^{(k+1)}|
\end{equation}
between $\mathbf{e}^{(k)}$ and $\mathbf{e}^{(k+1)}$, replacing the step using Lipschitz continuity in Proof~\ref{sec:proofmain}.
Following Proof~\ref{sec:proofmain}, $\mathbf{e}^{(l)}$ converges to $\mathbf{0}$:
\begin{equation}
\begin{split}
      \mathbf{0} & \leq |\mathbf{e}^{(l)}| \leq \frac{|\mathbf{M}|^{m+1}}{(m+1)!\epsilon^{m+1}}|\mathbf{e}^{(l+m+1)}| \\
    & \lim_{m\to\infty}\frac{|\mathbf{M}|^{m+1}}{(m+1)!\epsilon^{m+1}}|\mathbf{e}^{(l+m+1)}| = \mathbf{0}
    \end{split}
\end{equation}

\end{proof}
\hfill \qed

\begin{proof}
We also proof that for a fixed number of iterations $n$, $\alpha_l = $ or $\alpha_l = $ results in the same value when ReLU is used as activation function $\phi$. We separate the two versions into
\begin{equation}
    \mathbf{E}_{\beta}^{(l+1)} = \phi(\beta_l\mathbf{\Tilde{A}}\mathbf{E}_{\beta}^{(l)}\mathbf{W})
\end{equation}
with $\beta_l = \frac{1}{1+l\epsilon}$ and
\begin{equation}
     \mathbf{E}_{\gamma}^{(l+1)} = \phi(\gamma_l\mathbf{\Tilde{A}}\mathbf{E}_{\gamma}^{(l)}\mathbf{W})   
\end{equation}
with $\gamma_l = \frac{1}{1+(n-l-1)\epsilon}$ and show
\begin{equation}
    \mathbf{E}_{\beta}^{(n)} = \mathbf{E}_{\gamma}^{(n)}\, .
\end{equation}

The key fact we use is that 
\begin{equation}
    \alpha \phi(\mathbf{X}) = \phi(\alpha \mathbf{X})
\end{equation}
for any $\alpha > 0$ and $\phi$ as ReLU. Now we show that  
\begin{equation}
    |\mathbf{e}_{\beta}^{(n)} - \mathbf{e}_{\gamma}^{(n)}| = 0
\end{equation}
for the vectorized version following the same procedure as in previous proofs:
\begin{equation}
 \begin{split}
    |\mathbf{e}_{\beta}^{(n)} - \mathbf{e}_{\gamma}^{(n)}| &= |\phi(\beta_{n-1}\mathbf{M}\mathbf{e}_{\beta}^{(n-1)}) - \phi(\gamma_{n-1}\mathbf{M}\mathbf{e}_{\gamma}^{(n-1)})| \\
    & \leq |\mathbf{M}||\beta_{n-1}\mathbf{e}_{\beta}^{(n-1)} - \gamma_{n-1}\mathbf{e}_{\gamma}^{(n-1)}| \\
    & \leq |\mathbf{M}||\beta_{n-1}\phi(\beta_{n-2}\mathbf{M}\mathbf{e}_{\beta}^{(n-2)}) - \gamma_{n-1}\phi(\gamma_{n-2}\mathbf{M}\mathbf{e}_{\gamma}^{(n-2)})|\\
    & = |\mathbf{M}||\phi(\beta_{n-1}\beta_{n-2}\mathbf{M}\mathbf{e}_{\beta}^{(n-2)}) - \phi(\gamma_{n-1}\gamma_{n-2}\mathbf{M}\mathbf{e}_{\gamma}^{(n-2)})|\\
    & \leq |\mathbf{M}|^n|\beta_{n-1}\dots\beta_{0}\mathbf{e}_{\beta}^{(0)}) - \gamma_{n-1}\dots\gamma_{0}\mathbf{e}_{\gamma}^{(0)})| = 0
\end{split}   
\end{equation}
The last equality holds because $\beta_{n-1}\dots\beta_{0} = \gamma_{n-1}\dots\gamma_{0}$ and $\mathbf{e}^{(0)}_{\beta} = \mathbf{e}^{(0)}_{\gamma}$ is the initial state.

\end{proof}
\hfill \qed






\end{document}